\documentclass[preprint,12pt]{elsarticle}

\usepackage{amssymb}
\usepackage{hyperref}
\usepackage{multirow}
\usepackage{booktabs}
\usepackage{amsmath}
\usepackage{colortbl}
\usepackage{adjustbox}

\journal{Pattern Recognition Letters}

\begin{document}

\begin{frontmatter}



\title{Text as Environment: A Deep Reinforcement Learning Text Readability Assessment Model}


\author[inst1]{Hamid Mohammadi} %\corref{cor1}}

% \ead{hamid.mohammadi@aut.ac.ir}
% \cortext[cor1]{Corresponding author}

\author[inst2]{Seyed Hossein Khasteh}
\author[inst3]{Tahereh Firoozi}
\author[inst4]{Taha Samavati}

\affiliation[inst1]{organization={Computer Engineering Department, Amirkabir University of Technology},%Department and Organization
            city={Tehran},
            state={Tehran},
            country={Iran}}

\affiliation[inst2]{organization={Computer Engineering Department, K.N. Toosi University of Technology},%Department and Organization
            city={Tehran},
            state={Tehran},
            country={Iran}}

\affiliation[inst3]{organization={Measurement, Evaluation, and Data Science (MEDS), University of Alberta},%Department and Organization
            city={Edmonton},
            state={Alberta},
            country={Canada}}

\affiliation[inst4]{organization={School of Computer Engineering, University of Science and Technology},%Department and Organization
            city={Tehran},
            state={Tehran},
            country={Iran}}

\begin{abstract}
Evaluating the readability of a text can significantly facilitate the precise expression of information in written form. The formulation of text readability assessment involves the identification of meaningful properties of the text regardless of its length. Sophisticated features and models are used to evaluate the comprehensibility of texts accurately. Despite this, the problem of assessing texts' readability efficiently remains relatively untouched. The efficiency of state-of-the-art text readability assessment models can be further improved using deep reinforcement learning models. Using a hard attention-based \textbf{\textit{active inference}} technique, the proposed approach makes efficient use of input text and computational resources. Through the use of semi-supervised signals, the reinforcement learning model uses the minimum amount of text in order to determine text's readability. A comparison of the model on Weebit and Cambridge Exams with state-of-the-art models, such as the BERT text readability model, shows that it is capable of achieving state-of-the-art accuracy with a significantly smaller amount of input text than other models.
\end{abstract}

\begin{highlights}
\item Self-attention's computational load grows quadratically with input size
\item Transformers can be mathematically modeled as dynamic integral transforms
\item Domain decomposition method can reduce transformer's computational complexity
\item Reinforcement learning can be used for domain decomposition for transformers
\end{highlights}


\end{frontmatter}


\section{Introduction}
\label{Introduction}
% \linenumbers

{Text, as a prevalent form of communication, has a fundamental role in conveying knowledge and information between humans. Nevertheless, not all texts are equally intelligible and understandable for all people. It is, therefore, vital to measure the readability of written information in order to ensure its clarity and understandability. The significance of this measurement is apparent from its applications in different fields such as education \mbox{\cite{Rawian2019TextRA, cheng2022readability}}, medical instructions \mbox{\cite{devaraj2021paragraph, mac2022comparison, guo2021automated, zheng2022evaluation, ondov2022survey}}, social media communications \mbox{\cite{pancer2019readability, sazzed2022influence, gkikas2022text}}, marketing and advertising \mbox{\cite{curiel2021online, korniichuk2021conversion}}, and in some related fields of research like text simplification \mbox{\cite{murphy2022effect, monteiro2022using, cripwell2023document, ondov2022survey}}.}

As early as the 1940s, attempts were made to quantify the readability of text manually by reading experts. In order for such an evaluation to be standardized or globally accurate, researchers like \citet{flesch1943marks} have developed formulas for measuring the readability of texts. The readability formulas use simple and manually calculable properties of the text, such as the number of syllables, words, or sentences in the text, to assess its readability. Even today, these formulas have remained very popular and are still used by many people around the world.

Using heuristics and hand-made mathematical relations, the readability formulas are designed to assess the level of readability of texts in a particular language. As a result, they are usually low in accuracy and language-dependent. To compensate for the deficiencies of readability formulas, advanced and accurate readability assessment methods use machine learning techniques. Due to their use of NLP features and machine intelligence, these models can identify a proper level of readability based on extracted features. As such, these models represent a significant improvement over traditional readability formulas, providing much more accurate assessments of text complexity. Models proposed by \citet{vajjala2012improving}, \citet{xia2019text}, and \citet{mohammadi2018machine} are examples of state-of-the-art models for their target languages and target audience. In these models, Support Vector Machines (SVM) are trained on complex and extensive feature sets extracted from related datasets. However, due to language-specific NLP features, these models can be challenging to implement and are highly language-dependent. As a result, these models require significant pre-processing and language-specific NLP feature engineering to ensure accurate and successful implementation. This has led to the use of transformers \citet{vaswani2017attention} for natural language understanding and, therefore, text readability assessment. BERT is one of the most commonly used transformers in NLP applications, enabling state-of-the-art accuracy with minimal training data. Automating feature extraction with transformers has facilitated SOTA accuracy in various NLP applications and enabled multilingual models to be built efficiently.

It is not necessary to consider the entire length of the text when assessing its readability. For example, reading a fraction of a text can provide a general view of the text's readability for human assessors. It is surprising that previous approaches have not utilized this possibility to reduce computation load by efficiently processing the smallest amount of input. An improved variant of the BERT model is presented here that is based on reinforcement learning. This model can predict the readability of text using less than half the text's length. A reinforcement learning model is combined with a pre-trained BERT in this model to form \textbf{\textit{active inference}}. Input text can be perceived within a window of several adjacent words by the model. To view further parts of the text, the model's actions could move the hard attention window. As a result of the model's ability to intelligently choose which portion of the text is to be perceived, it is possible to determine the minimal amount of content to be read to determine the readability of the text.

The structure of the paper is as follows: Section \ref{Related Works} discusses the previous attempts to automate the readability assessment task in detail. Section \ref{Proposed Approach} presents the proposed model and describes its architecture. Later, section \ref{sec:experiments} reviews the experiment results, and section \ref{sec:discussion} explains the advantages and disadvantages of the presented model. The last section states the main contributions of this study and potential future works.

\section{Related Works}
\label{Related Works}
% \linenumbers

The text readability literature can be divided into four main categories: traditional formulas, machine learning models with hand-crafted features, deep learning models, and transformer models. The increasing complexity of models in each category improved the text readability models' accuracy, data efficiency, feature richness, and multi-lingual capabilities.

Flesch-Kincaid grade level \cite{flesch1943marks} can be named as one of the English language's earliest and most utilized readability formulas. The Flesch-Kincaid readability formula uses only the average number of words per sentence and the average number of syllables per word to evaluate text readability. The Flesch-Kincaid formula can be seen in \autoref{eq:flesch}.

\begin{multline}
Flesch\text{-}Kincaid\ Grade Level = \\ 0.39 \cdot\frac{|words|}{|sentences|}\ +\ 11.8 \cdot\frac{|syllables|}{|words|}\ -\ 15.59
\label{eq:flesch}
\end{multline}

With advances in automated computations, readability assessment applications utilize extracted and computer-calculated features. Lexile \cite{stenner1996measuring} and the work of \citet{collins2005predicting} used word-frequency and language models, respectively. It has been found that statistical models can be utilized to improve the accuracy of text readability assessment models. Formulas using traditional methods are simple to implement and require limited computational resources. In spite of these advantages, these methods are low in accuracy and have a significant difference between the results and human judgments \cite{benjamin2012reconstructing, hartley2016time, petersen2009machine, crossley2011text}. Due to the fact that these formulas are specially designed for a particular language, they cannot be applied to assess the readability of texts in other languages. Short text applications are also incompatible with these formulas, which are prevalent on social media and the web these days \cite{pilan2014rule}.

Using machine learning models, researchers have created a more accurate and comprehensive system for assessing text readability that overcomes the shortcomings of traditional formulas. Assessment of text readability can be viewed either as a regression problem or as a classification problem. However, studies have shown increased accuracy and applicability of text readability assessment as a classification task \cite{feng2010comparison}. Among the primary advantages of machine learning models are their use of many features (naive or sophisticated) and their automated ability to learn how the features interact with readability levels, which makes them more popular than traditional formulas \cite{franccois2012nlp}. Models are only as good as the features they include, which makes choosing features crucial. Simple features, such as the average number of characters or syllables in words, the average number of words in sentences, the number of sentences in a text, and simple statistical language models were features used in early machine learning models for text readability assessment like works presented in \cite{petersen2009machine, schwarm2005reading}. The use of syntactical features \cite{kate2010learning}, and cohesive features \cite{sung2015constructing} also supported the realization of models with higher accuracy. However, by restricting machine learning models to predefined features, it is impossible to extract and use subtle features that were not designed by NLP experts. To address this limitation, recent works have proposed techniques for automatically extracting features from unstructured texts, enabling more accurate models and providing a more efficient way to extract knowledge from text.

"Attention is all you need" is the slogan of the next generation of natural language understanding models \cite{vaswani2017attention}. The BERT model, a prominent model within the transformer family, incorporates self-attention and unsupervised transfer learning to improve many NLP tasks \cite{devlin2018bert}. Transformer models are able to understand long-term temporal dependencies by using soft attention as encoder layers. In addition, the use of arbitrary tasks to pre-train transformers increases their zero-shot and few-shot learning abilities \cite{devlin2018bert}. As compared to previous approaches to text analysis, transformers offer an unattainable advantage over previous approaches as transformers could benefit from large corpora of unlabeled text for pre-training. Based on the results of this paper, pre-trained BERTs can be fine-tuned and adapted to text readability problems. Transformers, however, have a limited input size as one of their constraints. By utilizing truncation, pre-trained transformers are applied to texts of different sizes. In addition, truncating a long text to fit the transformer's input size limits the perceived information since part of the text must be removed. Furthermore, in practical GPU implementations of transformers, applying a model to texts of uneven size wastes computational resources since excess parameters still need to be stored for efficient computation. Zero-padding and feature concatenation are naive approaches to adding flexible input sizes to transformers. Further, it is difficult to determine how much of the text needs to be processed to extract relevant and sufficient data. This paper proposes a method of optimally processing the minimum length of text to capture the necessary information required for text readability assessment.

{Reinforcement learning can be considered a semi-supervised approach to machine learning. The ability to learn from partially labeled data makes reinforcement learning particularly useful for NLP. Hence, there is a trend in using reinforcement learning models in NLP tasks such as machine translation \mbox{\cite{ke2022english, kang2020dynamic, uc2023survey, lam2018reinforcement}}, sentence simplification \mbox{\cite{zhao2020semi}}, text summarization \mbox{\cite{li2020text, keneshloo2019deep, liang2020abstractive}}, dialogue generation \mbox{\cite{yang2020multitask, saleh2020hierarchical}}, question answering \mbox{\cite{godin2019learning, qiu2020stepwise, nakano2021webgpt}}, and text generation \mbox{\cite{de2021survey, iqbal2022survey}}. Moreover, deep reinforcement learning models can help to fuse the advantages of reinforcement learning and deep learning so as to produce more accurate and efficient models for NLP tasks. Specifically, deep reinforcement learning could be used to develop hard-attention abilities. By incorporating hard-attention abilities, deep reinforcement learning models would be able to effectively identify and exploit the most relevant information from a given dataset, resulting in improved performance on NLP tasks. The main drawback of soft attention is the fast-growing computational load that makes them inefficient when processing large sequences of text. In contrast, hard-attention models reduce the size and computations of their models by decreasing the amount of information they process during each step \mbox{\cite{shen2018reinforced}}. Through \textbf{\textit{active inference}}, these models are able to intelligently focus on specific parts of a text that carry more valuable information for their tasks. Despite some drawbacks of deep reinforcement learning models, such as training instability, these models can achieve higher efficiency in NLP tasks in comparison to soft-attention deep learning models.}

We utilize reinforcement learning to optimize transformer-based text processing for texts of varying lengths. This approach combines the power of transformers, BERT in particular, with the \textbf{\textit{active inference}} ability of reinforcement learning models. The following sections discuss the proposed model in detail and compare it with standard approaches using technical evaluations of the readability datasets.

\section{Proposed Approach}
\label{Proposed Approach}

% Explain the exponential growth of computation using the transformer attention

The self-attention mechanism computations grow exponentially with increasing input size \cite{cao2022understand}. To show the relationship between input size and computational load, transformers can be mathematically modeled as dynamic integral transforms \cite{cao2022understand}. \autoref{eq:attention} shows the formulation of the self-attention coefficient.

\begin{equation}
\label{eq:attention}
\resizebox{.6\hsize}{!}{%
$A_{i\cdot} = \frac{e^{(QK^{\intercal}/\sqrt{d})_{i\cdot }}}{\sum_{j=1}^{p}e^{(QK^{\intercal}/\sqrt{d})_{ij}}} := Softmax(q_{i}K^{\intercal}/\sqrt{d})$%
}
\end{equation} \\

Where $A_{i\cdot}$ is the attention coefficient computed for the token at the $i$th position considering a total of $p$ tokens, the $Q$ and $K$ are query and key values, and the $q_{i}$ is the query computed for the $j$th token. Finally, $d$ is the embedding dimensions. The time complexity of \autoref{eq:attention} is $O(n^4)$ \cite{cao2022understand}. The growth rate of the self-attention mechanism makes it costly to apply to lengthy sequences. An empirical evaluation of the growth of the BERT model's latency relative to the input sequence length is presented in \autoref{tab:benchmark}.

\begin{table}[!htbp]
\centering
\caption{BERT model latency with different input sequence lengths and batch sizes on an Intel Xeon Platinum 8275 CPU (48 cores/96 threads) and using Tensorflow framework \cite{huggingface}. }
\label{tab:benchmark}
\begin{tabular}{lcccc}
    \toprule
\textbf{Batch size} & \textbf{Seq. Length (token)} & \textbf{Latency (ms)}  \\
    \midrule
\multirow{2}{*}{\textbf{16}} & 512 & 5,747 \\ & 256 & 1,663 \\ & 128 & 701 \\ & 64 & 473 \\
\cmidrule(l){2-3}
\multirow{2}{*}{\textbf{32}} & 512 & 11,800 \\ & 256 & 3,765 \\ & 128 & 1,518 \\ & 64 & 747 \\
\cmidrule(l){2-3}
\multirow{2}{*}{\textbf{128}} & 512 & 41,311 \\ & 256 & 13,562 \\ & 128 & 6,513 \\ & 64 & 3,065 \\
    \bottomrule
\end{tabular}
\end{table}

The Domain Decomposition Method (DDM) is a method for simplifying and solving integral equations. Essentially, DDM is a method for dividing a vector space into sub-spaces so that solvers can calculate the solution to each sub-problem more quickly. DDM is similar to the divide and conquer concept in computational algorithms. To reduce the computational requirement for accurate text classification, this study applies a DDM-like solution of dividing the problem domain into manageable sub-problems. According to \cite{cao2022understand}, the self-attention layer can be roughly described using an integral transform formulation. The integral transform form for a self-attention layer in \autoref{eq:selfattention} produces a Fredholm-like formulation for self-attention based on a kernel approximation.

\begin{equation}
\label{eq:selfattention}
\resizebox{.6\hsize}{!}{%
$z(x) \approx \frac{1}{\sum_{l=1}^{d}(e^{q_{i}K^{\intercal}/\sqrt{d}})_{l}} \int_{\omega}^{} \tilde{\kappa}(x,x^{'})v(x^{'})d\mu(x^{'})$%
}
\end{equation}\\

% Explain how the hard-attention method is more efficient in comparison to the attention computation growth (domain decomposition method)

Where $\tilde{\kappa}(x,x^{'})$ is the kernel approximation of the self-attention coefficient computed for two input tokens, $v(x^{'})$ is the linear projection of the token, and $\mu(x^{'})$ is the Borel function which is required for the conversion of the self-attention formulation into a continuous space. In this study, a \textbf{R}einforcement learning-based \textbf{A}ctive \textbf{I}nference \textbf{T}ransformer (\textbf{RAIT}) model is proposed that uses DDM to reduce the computational cost of text classification using transformers. RAIT differs from the standard DDM method in two main ways. Firstly, in DDM, all sub-problems are solved and utilized in forming the global solution. However, this study uses machine learning optimization to avoid the computation of all sub-problems by selecting the minimal set of sub-domains required to solve the global problem. Secondly, in a standard DDM, the global solution is computed by combining the sub-domain solutions via an auxiliary coarse problem. Yet, in this study, the global combination of sub-solutions is achieved using a learned policy in RL agent training. By converting a problem into an expected reward formulation, a problem can be framed as a reinforcement learning problem. \autoref{eq:discounted_reward} illustrates the general framework for expected rewards.

\begin{equation}
\label{eq:discounted_reward}
\resizebox{.6\hsize}{!}{%
$G_{t} = R_{t+1} + \gamma R_{t+2} + \gamma^{2}R_{t+3} + \cdots = \sum_{k=0}^{\infty}\gamma^{k}R_{t+k+1}$%
}
\end{equation}\\

In \autoref{eq:discounted_reward}, $G_{t}$ is the cumulative discounted reward, $R_{t}$ is the reward observed at time-step $t$, and $\gamma$ is the discount value. You can assign rewards for changing your attention window and selecting a class. Using $G_{t}$ general modeling of the desired task output (here: maximum accuracy with minimal computation), the Bellman equation can present an abstract global aggregation function for the custom DDM. The modified DDM is formulated as a state-value estimation problem in \autoref{eq:state} and state-action value estimation problems in \autoref{eq:stateaction}.

\begin{equation}
\label{eq:state}
\resizebox{.6\hsize}{!}{%
    $\upsilon_{\pi}(s) = E_{\pi}\left[ \sum_{k=0}^{\infty} \gamma^{k}R_{t+k+1} | S_{t} = z(x_{t+k+1}) \right]$%
}
\end{equation}

\begin{equation}
\label{eq:stateaction}
\resizebox{.6\hsize}{!}{%
$q_{\pi}(s,a)=E_{\pi}\left[ \sum_{k=0}^{\infty} \gamma^{k}R_{t+k+1}|S_{t}=z(x_{t+k+1}),A_{t}=a) \right]$%
}
\end{equation}\\

The $z(x)$ function replaces the $S_{t}$, which is the state representation at the time-step $t$. In other words, to reduce the problem's computational complexity, it is converted into a partially observable environment using reinforcement learning-based hard attention. Thus, the global solution is computed by solving only a small number of sub-domain problems. Evaluations presented in this paper demonstrate that this method reduces computational complexity without affecting the model's end-to-end accuracy.

\subsection{Active Inference}

% Text crop

Text is observed by the RAIT model using a text crop function. Crop function coordinates are derived from their internal state. As a first step, the internal state is initialized by focusing on the beginning of the text. The crop function's coordinates are updated in the next steps according to the RL agent's actions. The crop function's length (number of words to keep) could also be modified to adjust the trade-off between computational cost and accuracy. The evaluations in \autoref{sec:experiments} show the relationship between crop size and the model's accuracy.

% BERT

The texts are encoded using a BERT encoder \cite{vaswani2017attention}, and the produced representations are fed into the RL agent during training. The size of the crop function determines BERT encoder embedding sequence length. As demonstrated by \autoref{eq:stateaction}, reducing the crop size reduces the computation required by the BERT encoder by reducing the number of words included in the self-attention correlation matrix. The BERT encoder produces a fixed-size vector of 768 dimensions. Therefore, the reinforcement learning agent's network architecture does not depend on crop size.

% RL Agent

The reinforcement learning agent learns to predict text's category while observing a minimal set of text-window representations. The deep reinforcement learning agent is trained end-to-end with the BERT encoder using Q learning loss. Consequently, the negative reward for changing the crop function coordinates and the positive reward for correctly predicting the text's category are propagated using the error gradient through the DRL agent and the BERT encoder. Due to the discrete nature of the problem's formulation and the data efficiency of off-policy methods, a DQN is chosen for this study.

% Actions: See more text or select a class to terminate the process

The agent's actions can be divided into two categories. During the "see more" action, the crop function coordinates are changed programmatically so that the next text window can be loaded. The next cycle of text encoding and agent decision-making begins with this action. The "select class" action, on the other hand, terminates the classification loop. This action, which consists of multiple output nodes corresponding to the number of possible classes, selects the text's category based on the text windows observed up to this point.

\subsection{Model Training}

% Reward shaping

\begin{figure}[!htbp]
\centering
        \includegraphics[width=0.55\textwidth]{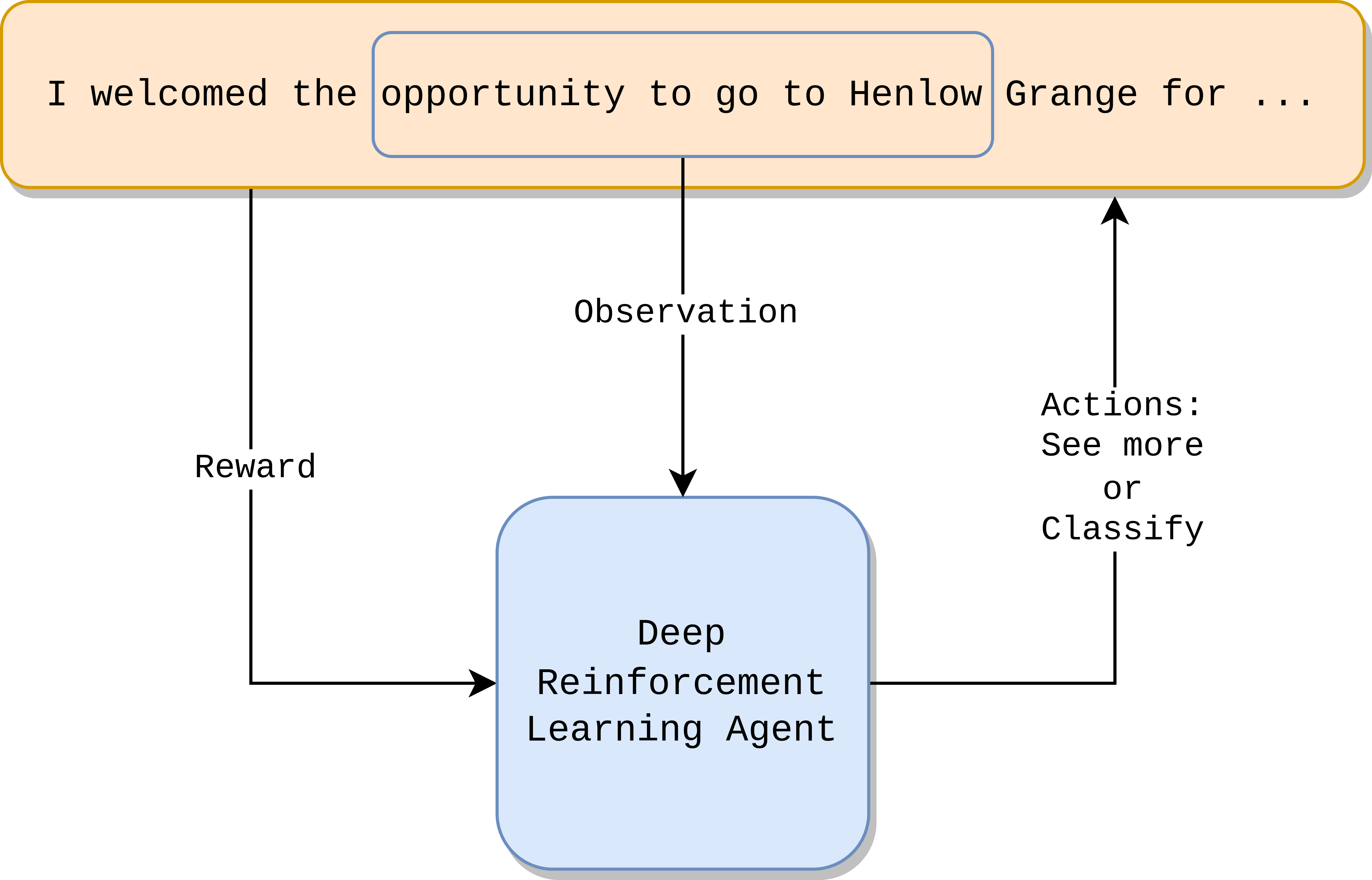}
    \caption{The interactions of the RAIT model with the text readability environment. BERT representations of words in the attention window act as the RL agent's observation. The agent decides to continue the observation or select the text's class based on observations and rewards.}
    \label{fig:interaction}
\end{figure}

The model is rewarded for its interactions with partially observable textual environments in two ways. Firstly, negative rewards (-0.1) are given for changes in window position to encourage the model to take the smallest number of steps. Secondly, the model can observe a positive (+1) or negative (-1) reward for choosing a readability level for the intended text by picking one of the readability classes. A visual depiction of model interaction with the textual environment is shown in \autoref{fig:interaction}.

\begin{figure}[!htbp]
\centering
        \includegraphics[width=0.97\textwidth]{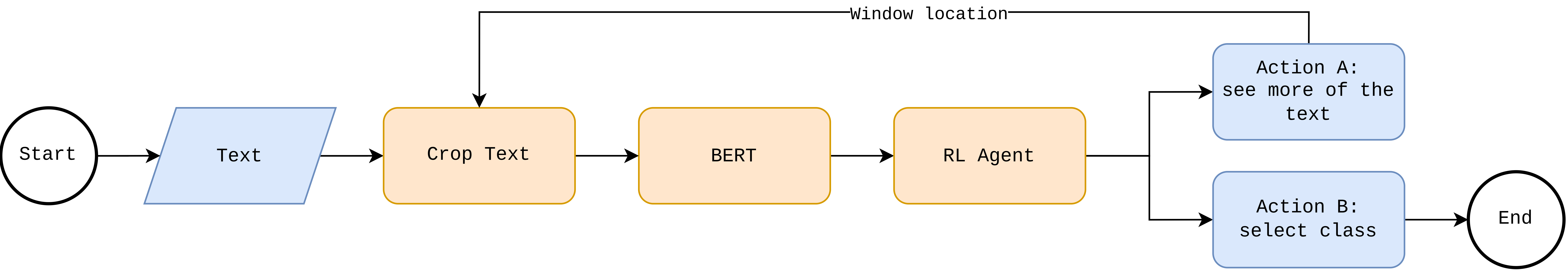}
    \caption{The classification process of the RAIT model. The crop text function is initialized to capture the left-most window at first. The RL agent's subsequent actions determine the crop window's future location. The processing loop is terminated when the RL agent selects a class for the input text.}
    \label{fig:drqn}
\end{figure}

% Exploration method

At each training step, the loss of the model is calculated using the Q learning equation (Equations \ref{eq:qlearning} and \ref{eq:loss}). Since deep reinforcement learning models are prone to divergence, a method called double Q network learning \cite{van2016deep} is applied to stabilize the process. For each state-action pair, the target Q value is computed using a frozen instance of the main model to avoid oscillations. As soon as the predefined training steps have been completed, the frozen network, which is called the target network, is replaced by a new copy of the main Q network.

\begin{equation}
\resizebox{.8\hsize}{!}
{
$Q(z(x_{t}),a_{t})=Q(z(x_{t}),a_{t})+\alpha\left[r_{t+1}+\gamma\underset{a}{\max}\: Q(z(x_{t+1}),a)-Q(s_{t}, a_{t})\right]$
}
\label{eq:qlearning}
\end{equation}

\begin{equation}
\resizebox{.6\hsize}{!}
{
$loss = \left(r_{t+1}+\gamma\underset{a}{\max}\: Q(z(x_{t+1}),a)-Q(z(x_{t}), a_{t})\right)^2$
}
\label{eq:loss}
\end{equation} \\

Deep reinforcement learning models demand numerous interactions with their environment to be sufficiently trained. To overcome this problem and further stabilize the learning process, a technique called experience replay \cite{lin1993reinforcement} is used in the current study. During its training process, the DRL model is fed multiple times with $(current-state, action, reward, next-state)$ tuples containing its previous interactions with the environment. The reintroduction of previous experiences to the model increases data usage efficiency and prevents the model from forgetting older experiences. Additional hyperparameters are presented in \autoref{tab:train}.

\begin{table}[!htbp]
\centering
\caption{Hyperparameters of the RAIT model. }
\label{tab:train}
\begin{tabular}{lc}
    \toprule
\textbf{Hyperparameter}     & \textbf{Value}           \\
    \midrule
Main network learning rate                 & $2\mathrm{e}{-5}$ \\
Number of training episodes		& $2.5\mathrm{e}{5}$ \\
Target network update frequency	& 5 episodes \\
Experience buffer size			& 1000 \\
Q learning discount factor			& 0.99 \\
Exploration policy			& $\epsilon-greedy$ \\
Initial action randomness			& 100\% \\
Final action randomness			& 10\% \\
    \bottomrule
\end{tabular}
\end{table}

\section{Experiments}
\label{sec:experiments}

% Explain how the experiments show the effectiveness of your method

Using \textit{\textbf{active inference}}, the proposed method effectively decreases the computational cost of transformer-based text readability assessment models. Mathematically, the suggested method reduces computational costs by utilizing a smaller self-attention perception field than a standard transformer. A comparison of the RAIT model with standard baselines in text readability assessment is presented in this section to demonstrate that it is effective at reducing the computing requirements of text readability assessment without significantly adversely affecting accuracy at the end-to-end level.

\subsection{Datasets}

RAIT is evaluated using two text readability datasets. Firstly, the Weebit dataset \cite{vajjala2012improving} is used to assess the model's accuracy in deciding the readability of English texts for native readers. Weebit is gathered from articles in the Weekly Reader magazine and BBC-Bitesize, which are targeted at readers of different ages. There are five different readability levels in the Weebit dataset arranged by age (8-9, 9-10, 10-12, 12-14, 14-16). More than ten thousand texts are included in the dataset. In addition to the Weebit dataset, RAIT is applied to the Cambridge Exams dataset \cite{xia2019text} in order to assess its English as a second language readability. This dataset's text comes from the Cambridge English Exams reading section, which targets students at five CEFR levels (A2 to C2). In comparison to the Weebit dataset, the Cambridge dataset contains only 331 texts. Details about the datasets can be found on \autoref{tab:datasets}.

\begin{table}[!htbp]
\centering
\caption{{Number of texts and length of texts at each readability level before balancing.}}
\label{tab:datasets}
\begin{tabular}{lccc}
    \toprule
\textbf{Dataset} & \textbf{Class} & \textbf{\# texts} & \textbf{Avg. \# words/text} \\
    \midrule
\multirow{1}{*}{\textbf{Weebit} \cite{vajjala2012improving}}
& 8-9 & 629 & 150 \\
& 9-10 & 789 & 189 \\
& 10-12 & 807 & 289 \\
& 12-14 & 646 & 238 \\
& 14-16 & 7615 & 347 \\
& total & 10486 & 311 \\
\cmidrule(l){2-4}
\multirow{1}{*}{\textbf{Cambridge} \cite{xia2019text}}
& A2 & 64 & 139 \\
& B1 & 60 & 268 \\
& B2 & 71 & 613 \\
& C1 & 67 & 768 \\
& C2 & 69 & 752 \\
& total & 331 & 519 \\
    \bottomrule
\end{tabular}
\end{table}

% Talk about the balance in the datasets
{According to Table {\ref{tab:datasets}}, the Cambridge dataset is fairly balanced, while the Weebit dataset has extreme imbalance in its fifth class (14-16). A large difference in data point counts between classes can lead to imbalanced learning problems. Therefore, similarly to Weebit's original paper {\cite{vajjala2012improving}}, a total of 3145 texts are used for evaluation purposes.}

\subsection{Baselines}

Common baselines and state-of-the-art models are implemented to evaluate the presented model comprehensively. Transformers, led by the BERT model, have dominated natural language processing. A comparison between RAIT and the BERT model evaluates the accuracy of the method when compared to the current state-of-the-art in-text readability models. Moreover, Word2Vec-based ConvLSTM models act as their low-computation, low-accuracy counterparts to demonstrate the superior accuracy of the proposed method at a low computation cost.

\subsubsection{Convolutional-LSTM}

The Word2Vec ConvLSTM \cite{hassan2017deep} is a sequence classification model. The vector space representation of words in the input text window is combined to form a 2-dimensional view of the input text (1-dimensional Word2Vec $\times$ 1-dimensional text). In the ConvLSTM model, convolutional layers act as feature extraction modules, encoding information as a window embedding from the input. In order to create a text embedding, the window embeddings are aggregated using LSTM layers. Therefore, ConvLSTM operates as a Text2Vec model. A supervised cross-entropy loss feeds the convolutional backbone and LSTM layers with their learning signals. The architecture of the ConvLSTM-based text readability assessment model can be found in \autoref{fig:conv_lstm_arch}.

\begin{figure}[!htbp]
\centering
        \includegraphics[width=0.97\textwidth]{ConvLSTM.drawio.png}
    \caption{The architecture of the ConvLSTM model. The Word2Vec representation sequence is fed to the convolutional backbone for feature extraction. With the help of an LSTM layer, the extracted features for each word are aggregated into a text-level feature. Finally, an MLP classifies the aggregated features according to their readability.}
    \label{fig:conv_lstm_arch}
\end{figure}

\subsubsection{BERT}

Considering the BERT model's unprecedented and excellent results in several NLP tasks, it is exciting to see BERT's accuracy on the text readability assessment task. The BERT model was trained on a large-scale corpus, and its deep learning architecture allows it to capture the subtle nuances of language. Thus, it is expected to provide highly accurate readability assessments, as it can detect and model the complexities of a text. The implemented BERT-based text readability assessment model works by classifying the text representations generated by a fine-tuned BERT model from the texts in each dataset \cite{devlin2018bert}. Similarly, RAIT also relies on the BERT model as a backbone, but the fine-tuning process is done by the RL loss and through the agent's interaction with the textual environment. In this study, the BERT base model is used, which has approximately 110 million parameters \cite{devlin2018bert}.

\subsection{Results}

Datasets are divided into two parts, 80 percent as the training dataset and 20 percent as the testing dataset. The datasets are bootstrapped to ensure their validity and stability. Comparative analyses are conducted with different window sizes applied to the models in order to demonstrate the model's ability to accurately and efficiently predict text readability. Benchmark models also use limited window sizes of 256, 128, and 64 words in addition to the full text as input. Each model's perception is therefore limited to the first N words in a text. The defined window sizes are used in training and evaluating all three models. We train models with different hyperparameters, and we report the results that are obtained with the best hyperparameters. {Table {\ref{tab:models}} compares some of the recently published models for English text readability on the Weebit {\cite{vajjala2012improving}} and Cambridge {\cite{xia2019text}} readability datasets with BERT.}

\begin{table}[!htbp]
\centering
\caption{{Model accuracy comparison on the Weebit {\cite{vajjala2012improving}} and Cambridge {\cite{xia2019text}} readability datasets.}}
\label{tab:models}
\begin{adjustbox}{width=\textwidth}
\begin{tabular}{lcccl}
\toprule
\textbf{Model} &
\textbf{Weebit \cite{vajjala2012improving}} &
\textbf{Cambridge \cite{xia2019text}} &
\textbf{Design} \\
\midrule
\citet{vajjala2012improving} & 93.3\% & - & SML (SVM) + hand-crafted features \\
ConvLSTM \cite{hassan2017deep} & 74.4\% & 76.3\% & Convolutional NN + LSTM \\
\citet{xia2019text} & - & \textbf{80.3\%} & SML (SVM) + hand-crafted features \\
\citet{fujinuma2021semi} & - & 79.6\% & Graph Convolutional Network \\
\citet{qiu2021learning} & 87.3\% & 78.5\% &  Graph NN + Transformer \\
\citet{li2022unified} & 92.7\% & - & Transformer + hand-crafter features \\
\citet{jian2022english} & 89.1\% & - & Convolutional NN + Pooling \\
BERT \cite{devlin2018bert} & \textbf{94.3\%} & 76.1\% & Transformer \\
\bottomrule
\end{tabular}
\end{adjustbox}
\end{table}

% Compare your results with other SOTA models
{Table \mbox{\ref{tab:models}} demonstrates BERT's superior accuracy (94.3\%) on the Weebit dataset. The second-best accuracy on this dataset is achieved by the SVM model proposed by \mbox{\citet{vajjala2012improving}} which uses hand-crafted features. Additionally, the original model proposed by \mbox{\citet{xia2019text}} achived state-of-the-art accuracy (80.3\%) on the Cambridge dataset. However,  SVM-based models require the design and implementation of hand-crafted features which is costly, and limits the generalizability of the model across languages and tasks. As a result, despite the speed and accuracy of traditional models, deep learning- and transformer-based models are more widely used because of their feature extraction automation. Automated feature extraction reduces task- and language-specific biases in the model which in turn increases the need for more data. The data requirement especially affects deep models' performance on datasets of smaller sizes (i.e., Cambridge). Use of small and biased models, more training data, and transfer learning is preferred in these scenarios.}

{The RAIT model increases the computational efficiency of the BERT model by reducing self-attention computations inside the transformer backbone. The comparison between the proposed model results and the other state-of-the-art models is shown in \mbox{\autoref{tab:res-com}}. According to the reported accuracy and latency, a reduction in the input length of the BERT model reduces its latency while negatively affecting its accuracy. On the other hand, the RAIT model can avoid drastic accuracy loss by actively adjusting the length of text required for accurate classification. As a result, the RAIT model can achieve close to state-of-the-art accuracy with less than half of the original BERT's latency. For example, the RAIT models with a window size of 64 achieve an accuracy of 93.8\% on the Weebit dataset with a latency of 2.1 ms. In other works, the 83\% reduction in computational cost with only 0.5\% loss in accuracy.}

\begin{table}[!htbp]
\centering
\caption{{Comparison between the proposed model and the state-of-the-art models on different datasets. The execution times were measured using the Huggingface {\cite{huggingface}} Pytorch {\cite{paszke2017automatic}} implementation of BERT {\cite{devlin2018bert}} on an A100 GPU with a batch size of 1.}}
\label{tab:res-com}
\noindent\begin{tabular}{lccc}
\toprule
\textbf{Model (window-size)} &
\textbf{Weebit \cite{vajjala2012improving}} &
\textbf{Cambridge \cite{xia2019text}} &
\textbf{Latency (ms)} \\
\midrule
BERT (full) & \textbf{0.943} & 0.761 & 12.6 \\
BERT (256) & 0.916 & 0.734 & 3.8 \\
BERT (128) & 0.911 & 0.729 & 1.6 \\
BERT (64) & 0.887 & 0.712 & 1.0 \\
\cmidrule(l){2-4}
RAIT (full) & \textbf{0.943} & 0.762 & 12.8 \\
RAIT (256) & \textbf{0.942} & \textbf{0.76} & 4.6 \\
RAIT (128) & \textbf{0.939} & \textbf{0.758} & 2.7 \\
RAIT (64) & \textbf{0.938} & \textbf{0.755} & 2.1 \\
\bottomrule
\end{tabular}
\end{table}

\section{Discussion}
\label{sec:discussion}

% Explain the results in detail
%% Explain the average move on each of the window sizes
Based on evaluation results, RAIT maintains state-of-the-art accuracy while significantly reducing model size and computation load. Similar to its original BERT counterpart, the full window-size model behaves like a standard transformer. The "see more" action is virtually eliminated when the window size is at its maximum. As a result, the proposed model differs only in its reinforcement learning loss instead of its multiclass cross-entropy loss from the BERT model in this scenario. However, the presented method is more accurate when the window size is limited to a value less than the maximum possible size. Based on the proposed method results, the 256, 128, and 64-word window sizes achieve state-of-the-art accuracy while observing 1.2, 1.7, and 2.1 windows of words, respectively. Conversely, models such as BERT and ConvLSTM show reduced accuracy when limited to smaller observation windows.

% Advantages of the proposed method
%% Insignificant accuracy loss with lower computational cost
%% Higher accuracy in comparison to the same input length model
\textit{\textbf{Active inference}} allows the proposed method to accurately and efficiently classify partially observed input data based on the model's representation. In other words, a dynamic selection of observed windows is used to maximize the accuracy of each input text and minimize the model's load. It is reasonable to use dynamic observation lengths due to the fact that different texts' representations have varying marginal errors within their class cluster. Text representations closer to class cluster boundaries must be more accurate to avoid misclassification. RAIT can also adjust the global solution aggregation strategy to maintain accuracy close to the state-of-the-art. When the window size is smaller, the self-attention function is further limited to generating contextually relevant text representations. However, as presented in \autoref{tab:res-com}, the proposed method adapts to the smaller window size by increasing the number of observed windows. The limited self-attention perception range negatively impacts the quality of learned representations, but the reinforcement learning-based global solution aggregation described in Equations \ref{eq:state} and \ref{eq:stateaction} minimizes the negative impact when compared to state-of-the-art models.

\section{Conclusion and future work}
\label{sec:con}
% \linenumbers

% Conclusion

The readability of a text can be determined precisely by analyzing a small section of it. This study introduces a reinforcement learning-based hard attention method based on transformers in order to realize this point. This hard attention method allows for precise readability assessment by focusing on the most informative parts of the text. RAIT uses the Domain Decomposition Method (DDM) to divide the computationally expensive attention mechanism across smaller and more manageable portions of the text. DDM further reduces computation complexity by calculating the global solution from a minimal set of sub-problem solutions using reinforcement learning. As evidenced by the comparison between the newly developed method and standard baselines, computational cost reduction has minimal effect on model accuracy.

The next step in this research is to test the RAIT model on additional NLP tasks, including automated essay scoring, sentiment analysis, and text classification in general. This will assess the generality of this paper's assumptions. Combining this method with other deep computation reduction methods like distillation can further optimize transformer-based text processing methods. Currently, the proposed method does not interact with the input text beyond simple actions by the reinforcement learning agent. There is no possibility for the agent to change the observation window size dynamically or jump to different text sections randomly. The global solution aggregation strategy can be further optimized with these actions.

\section*{Acknowledgements}
% \linenumbers

We would like to express my deepest appreciation to Dr. Ekaterina Kochmar\footnote{Department of Computer Science and Technology, University of Cambridge}, Dr. Baosong Yang\footnote{Department of Computer and Information Science, University of Macau}, and Dr. Yang Gao\footnote{Department of Computer Science, Royal Holloway, University of London} for their valuable review and suggestions on this manuscript. We are also grateful to HuggingFace \cite{huggingface} and StableBaselines3 \cite{stable-baselines3} for providing us with state-of-the-art tools and frameworks for natural language processing and reinforcement learning. Their open-source contributions have greatly facilitated our research and experiments.

 \bibliographystyle{elsarticle-num-names} 
 \bibliography{cas-refs}





% \end{thebibliography}
\end{document}